\documentclass{article}
\usepackage{ijcai20}

\usepackage{times}
\usepackage{soul}
\usepackage{url}
\usepackage[hidelinks]{hyperref}
\usepackage[utf8]{inputenc}
\usepackage[small]{caption}
\usepackage{graphicx}
\usepackage{amsmath}
\usepackage{amsthm}
\usepackage{booktabs}
\usepackage{algorithm}
\usepackage{algorithmic}
\usepackage{multirow}
\usepackage{amssymb}
\usepackage{bm}

\title{Object-Aware Multi-Branch Relation Networks for Spatio-Temporal Video Grounding}

\author{
Zhu Zhang$^1$
\and
Zhou Zhao$^{1}$\footnote{Zhou Zhao is the corresponding author.} \and
Zhijie Lin$^1$\and
Baoxing Huai$^2$\And
Jing Yuan$^2$
\affiliations
$^1$College of Computer Science, Zhejiang University, China\\
$^2$Huawei Cloud \& AI, China
\emails
\{zhangzhu, zhaozhou, linzhijie\}@zju.edu.cn,
\{huaibaoxing, nicholas.yuan\}@huawei.com
}

\begin{document}

\maketitle

\begin{abstract}
Spatio-temporal video grounding aims to retrieve the spatio-temporal tube of a queried object according to the given sentence. Currently, most existing grounding methods are restricted to well-aligned segment-sentence pairs. In this paper, we explore spatio-temporal video grounding on unaligned data and multi-form sentences. This challenging task requires to capture critical object relations to identify the queried target. However, existing approaches cannot distinguish notable objects and remain in ineffective relation modeling between unnecessary objects. Thus, we propose a novel object-aware multi-branch relation network for object-aware relation discovery. Concretely, we first devise multiple branches to develop object-aware region modeling, where each branch focuses on a crucial object mentioned in the sentence. We then propose multi-branch relation reasoning to capture critical object relationships between the main branch and auxiliary branches. Moreover, we apply a diversity loss to make each branch only pay attention to its corresponding object and boost multi-branch learning. The extensive experiments show the effectiveness of our proposed method.
\end{abstract}

\section{Introduction}
\begin{figure}[t]
\centering
\includegraphics[width=0.45\textwidth]{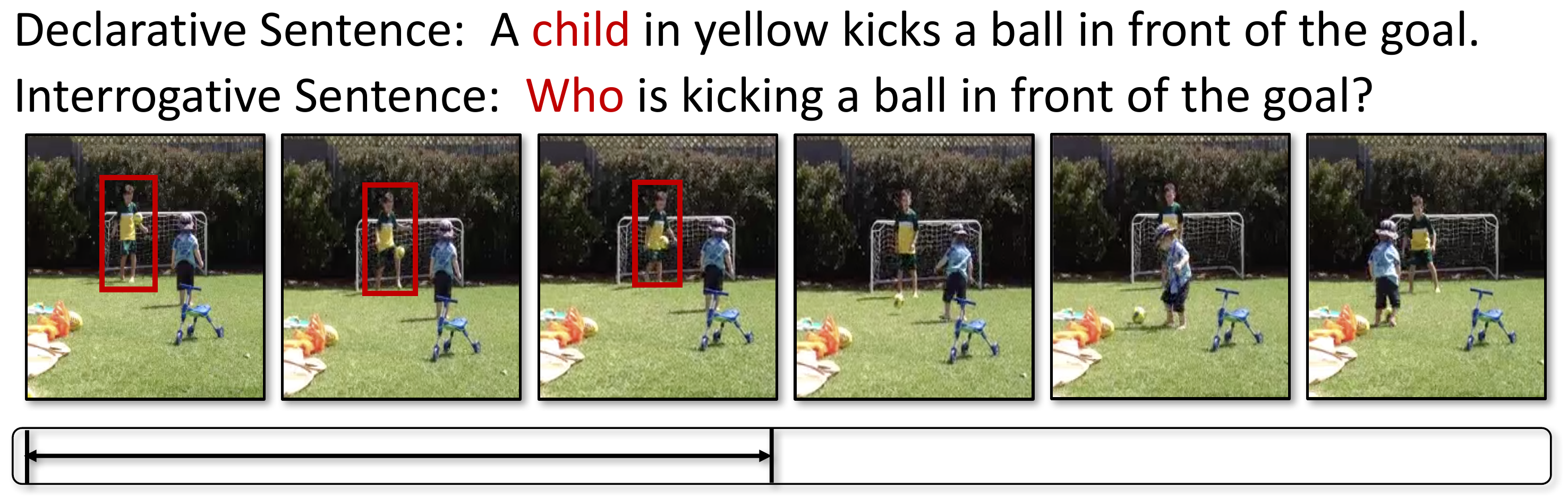}
\caption{An example of spatio-temporal video grounding on unaligned data and multi-form sentences.}\label{fig:exp}
\end{figure}

Spatio-temporal video grounding is an emerging task in the cross-modal understanding field.
Given a sentence depicting an object, this task aims to retrieve the spatio-temporal tube of the queried object, i.e. a sequence of bounding boxes. Most existing spatio-temporal grounding methods~\cite{chen2019weakly,shi2019not} are restricted to well-aligned segment-sentence pairs, where the segment has been trimmed from the raw video and is temporally aligned to the sentence. 
Recently, researchers~\cite{zhang2020does} begin to explore spatio-temporal video grounding~(STVG) on unaligned data and multi-form sentences. 
Concretely, as shown in Figure~\ref{fig:exp}, either the declarative sentence or interrogative sentence describes a short-term action of the target object "child" and matches with the spatio-temporal tube within a small segment. 
To localize the target existence in a fleeting clip, we need to distinguish the subtle status of the object according to the textual information. 
Specifically, the sentence often illustrates diverse relationships between the queried object with other objects, thus the key of this task is to capture these crucial relations in video contents to identify the spatio-temporal tube.
Particularly, the interrogative sentences depict unknown objects and lack the explicit information of the object, e.g., the direct characteristics "a child in yellow" in Figure~\ref{fig:exp}. Grounding these sentences can only depend on the object relationships such as the action relation "kicking a ball" and spatial relation "in front of the goal". 

Although existing grounding methods~\cite{yamaguchi2017spatio,chen2019weakly} have achieved excellent performance on aligned segment-video pairs, they are ineffectively applied to unaligned data and multi-form sentences. On the one hand, they heavily rely on the tube pre-generation to extract a series of candidate tubes and then select the most relevant one according to the sentence. But without the temporal alignment, it is difficult to pre-generate reasonable candidate tubes. On the other hand, these approaches always ignore the relation construction between objects and model each tube individually.
Recently, Zhang et al.~\cite{zhang2020does} explore spatio-temporal grounding on unaligned data. 
They incorporate the textual clues into region features and employ spatio-temporal graph reasoning to retrieve the spatio-temporal tube. Despite this method tries to capture object relations by cross-modal region interactions, it does not exclude inessential regions from massive region proposals and may lead to severe obstruction for effective relation modeling.
Concretely, there are a large number of objects in videos but most of them are irrelevant to the textual descriptions.
~\cite{zhang2020does} fails to filter out the unnecessary ones and remain in the coarse relation modeling for all regions.
Hence, we need to pay more attention to the crucial objects mentioned in the sentence and build sufficient cross-modal relation reasoning between them for precise video grounding.

In this paper, we propose a novel Object-Aware Multi-Branch Relation Network~(OMRN) for object-aware fine-grained relation reasoning. 
We first extract region features from the video and learn object representations corresponding to the nouns in the sentence.  We then employ multiple branches to develop object-aware region modeling and discover the notable regions containing informative objects, where the main branch corresponds to the queried object and each auxiliary branch focuses on an object mentioned in the sentence. 
Concretely, we apply the object-aware modulation to strengthen object-relevant region features and weaken unnecessary ones in each branch.  
Next, we can conduct object-region cross-modal matching in each branch. After it, we propose the multi-branch relation reasoning to capture critical object relationships between the main branch and auxiliary branches, where the irrelevant regions are filtered out by preceding matching scores. 
Further, considering each branch should only focus on its corresponding object, we devise a diversity loss to make different branches pay attention to different regions, that is, have diverse distributions of matching scores. 
Eventually, we apply a spatio-temporal localizer to determine the temporal boundaries and retrieve the target tube.
The main contributions of this paper are as follows:
\begin{itemize}
\item We propose a novel object-aware multi-branch relation network for spatio-temporal video grounding, which can discover object-aware fine-grained relations and retrieve the accurate tubes of the queried objects.
\item We devise multiple branches with a diversity loss to develop object-aware region modeling, where each branch focuses on a crucial object mentioned in the sentence and the diversity loss makes different branches focus on their corresponding objects. 
\item We employ the multi-branch relation reasoning to capture critical object relationships between the main branch and auxiliary branches. 
\item The extensive experiments show the effectiveness of our object-aware multi-branch framework.
\end{itemize}

\section{Related Work}
In this section, we briefly review some related work on visual grounding and video grounding.

Visual grounding is to localize the object in an image according to the referring expression. Early approaches~\cite{hu2016natural,nagaraja2016modeling} often model the language information by RNN, extract region features through CNN and then learn the object-language alignment. 
Recent methods~\cite{yu2018mattnet,hu2017modeling} parse the expression into multiple parts and compute cross-modal alignment scores for each part.
Furthermore, \cite{deng2018visual,zhuang2018parallel} employ the co-attention mechanism to develop cross-modal interactions for fine-grained matching. 
And~\cite{yang2019dynamic,yang2019cross} capture the relations between regions to boost the grounding accuracy.

Video grounding can be categorized into temporal grounding and spatio-temporal grounding.
Given a sentence, temporal grounding localizes a temporal clip in the video.
Early methods~\cite{hendricks2017localizing,gao2017tall} apply a proposal-and-selection framework that first samples massive candidate clips and then select the most relevant one semantically matching with the sentence. 
Recently, \cite{chen2019localizing,zhang2019cross,lin2020moment} develop frame-by-word interactions between visual and textual contents and discover the dynamical clues by attention mechanism.
\cite{zhang2019man} explicitly model moment-wise relations as a structured graph and employ an iterative graph adjustment.
\cite{yuan2019semantic} propose a semantic conditioned dynamic modulation for better correlating video contents over time and \cite{zhang2019learning} adopt a 2D temporal map to cover diverse moments with different lengths. 
In the weakly-supervised setting,
\cite{mithun2019weakly} leverage the attention scores to align moments with sentences, and \cite{lin2019weakly} propose a semantic completion network to estimate each clip by language reconstruction.  
Besides natural language queries, Zhang et al.~\cite{zhang2019localizing} try to detect the unseen video clip according to image queries.

Spatio-temporal video grounding is a natural extension of temporal grounding, which retrieves a spatio-temporal tube from a video corresponding to the sentence. Most existing approaches are designed for well-aligned segment-sentence data. 
\cite{yamaguchi2017spatio} only ground the person tube in multiple videos and \cite{zhou2018weakly,chen2019weakly} further retrieve the spatio-temporal tubes of diverse objects from trimmed videos by weakly-supervised MIL methods.
Different from single-object grounding, \cite{Huang_2018_CVPR,shi2019not} localize each noun or pronoun of sentences in frames. 
Recently, \cite{zhang2020does} explore spatio-temporal grounding on unaligned data by spatio-temporal cross-modal graph modeling. But it still fails to capture the critical objects and remains in the coarse relation reasoning. 
In this paper, we explore object-aware fine-grained relation modeling and further improve the grounding accuracy.

\begin{figure*}[t]
\centering
\includegraphics[width=0.9\textwidth]{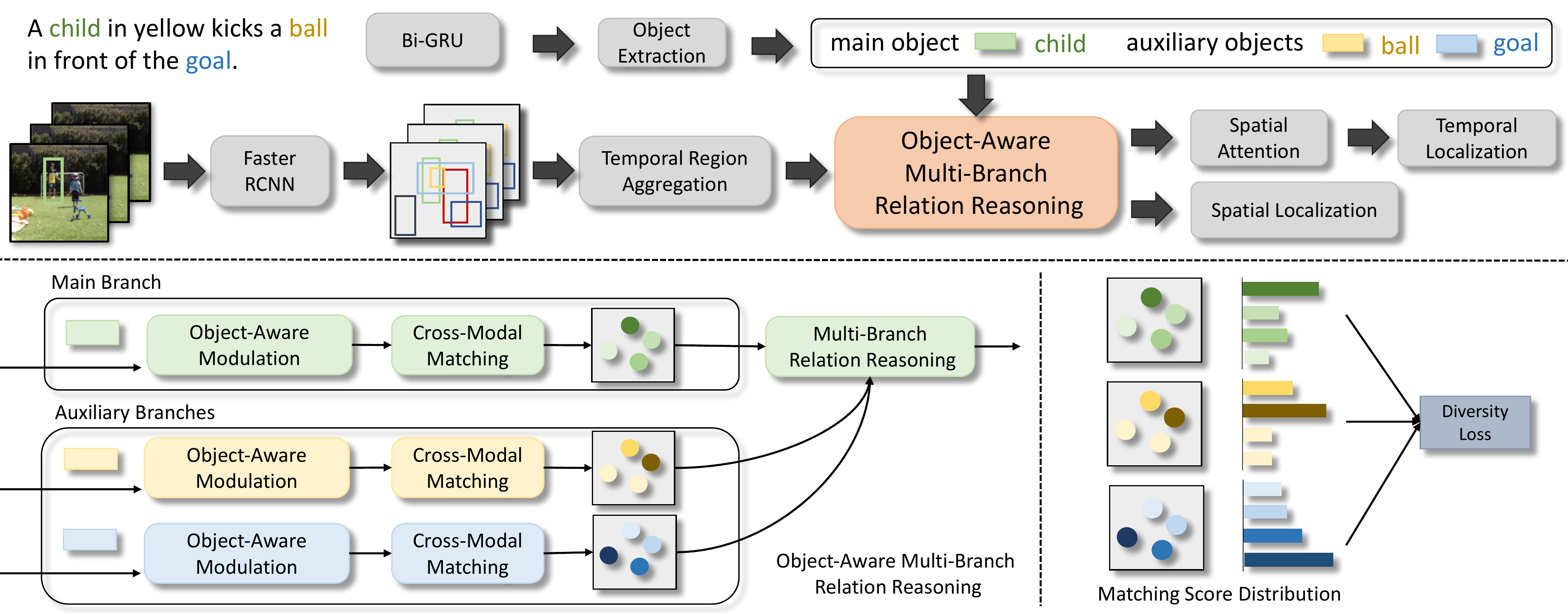}
\caption{The Overall Architecture of Object-Aware Multi-Branch Relation Network.
 }\label{fig:framework}
\end{figure*}

\section{The Proposed Method}
As shown in Figure~\ref{fig:framework}, we propose the object-aware multi-branch relation network (OMRN) for this STVG task, where we develop object-aware multi-branch region modeling to discover the notable regions containing informative objects and devise multi-branch relation reasoning to capture critical object relationships.

\subsection{Region and Object Extraction}
Given a video  ${\bf v}$, we extract region features $\{ \{{\bf r}^{n}_k\}_{k=1}^K \}_{n=1}^N$ by a pre-trained Faster R-CNN, where the video has $N$ frames and each frame contains $K$ regions. The feature ${\bf r}^{n}_k$ corresponds to the $k$-th region in frame $n$. And each region is also associated with a bounding box ${\bf b}^n_k = (x^n_k, y^n_k,w^n_k,h^n_k)$, where $(x^n_k, y^n_k)$ are the center coordinates and $(w^n_k,h^n_k)$ are the box width and height.
Considering video grounding requires to capture the object dynamics but region features extracted from still frames lack motion information, we then adopt a temporal region aggregation method to incorporate dynamic clues from forward $L$ frames and backward $L$ frames into each region. Concretely, if two regions in adjacent frames have similar semantic features and spatial locations, we expect them to contain the same object. Thus, the linking score between region ${\bf r}^{n_1}_{i}$ and ${\bf r}^{n_2}_{j}$ by
\begin{eqnarray}
& s({\bf r}_i^{n_1}, {\bf r}_j^{n_2}) = {\rm cos}({\bf r}_i^{n_1}, {\bf r}_j^{n_2}) + \frac{\alpha}{|n_2-n_1|} \cdot {\rm IoU}({\bf b}_i^{n_1}, {\bf b}_j^{n_2}),  \nonumber
\end{eqnarray}
where the ${\rm cos}(\cdot)$ means the cosine similarity of two features and ${\rm IoU}(\cdot)$ is the IoU score of their bounding boxes. The $|n_2-n_1|$ is the temporal distance of two regions and applied to limit the IoU score. And $\alpha$ is a balanced coefficient. For each region, we select the region with the maximal linking score from each adjacent frame and obtain $2L$ relevant features. Next, we apply a mean pooling on the $2L+1$ features to absorb region dynamics, followed by a linear layer for feature transformation. For simplicity, we still denote the pooled region features with temporal dynamics by $\{ \{{\bf r}^{n}_k\}_{k=1}^K \}_{n=1}^N$.

For language modeling, we input word embeddings of the sentence into a Bi-GRU to learn the word-level semantic features $\{{\bf s}_m\}_{m=1}^M$ with $M$ words. 
After it, we learn object features with context information for each object mentioned in the sentence. Specifically, we first identify all nouns in the sentence using the library of NLTK.  Assuming there are $T$ nouns in the sentence, each noun points to an object in the video and the $t$-th noun corresponds to the word-level feature ${\bf s}_t$. In Figure~\ref{fig:framework}, the sentence contains three objects "child", "ball" and "goal", and we aim to retrieve the spatio-temporal tube of the main object "child" (i.e. the queried object). For interrogative sentences, we regard the interrogative words (e.g. "who" and "what") as the main objects.
Next, we apply a context attention to aggregate the language context for each object by
\begin{eqnarray}
&\beta_{t,m} = {\bf w}^{\top}{\rm tanh}({\bf W}_1^s {\bf s}_{t} + {\bf W}_2^s{\bf s}_{m} + {\bf b}^s),  \nonumber\\
&{\bf \widetilde o}_t = \sum \limits_{m=1}^{M} {\rm softmax}({\beta}_{t,m})\cdot {\bf s}_{m}, \ {\bf o}_t = [{\bf s}_t;{\bf \widetilde o}_t], \nonumber
\end{eqnarray}
where ${\bf W}_1^s$, ${\bf W}_2^s$ are projection matrices, ${\bf b}^s$ is the bias and ${\bf w}^{\top}$ is the row vector. The ${\beta}_{t, m}$ is the attention weight of object $t$ for the $m$-th word. Finally, we obtain the object features $\{{\bf o}_t\}_{t=1}^T$, where ${\bf o}_1$ is the feature of the main object and $\{{\bf o}_t\}_{t=2}^T$ are auxiliary object features.

\subsection{Object-Aware Multi-Branch Relation Reasoning}
We next devise object-aware multi-branch relation reasoning with a diversity loss to capture object-aware fine-grained relations. Concretely, we first take multiple branches to learn object-aware region features and then apply multi-branch relation reasoning to capture critical object relationships from multiple branches, where the main branch corresponds to the main object and auxiliary branches correspond to auxiliary objects. And the diversity loss makes different branches focus on their corresponding objects. 

\subsubsection{Object-Aware Multi-Branch Region Modeling}
For branch $t$ with object ${\bf o}_t$, we first apply the object-aware modulation to strengthen object-relevant region features and weaken unnecessary ones. Concretely, we produce the object-aware modulation vectors by 
\begin{eqnarray}
& {\bm \gamma}_t =  {\rm tanh}({\bf W}^{\gamma} {\bf o}_{t}+ {\bf b}^{\gamma}), \ {\bm \delta}_t = {\rm tanh}({\bf W}^{\delta} {\bf o}_{t}+ {\bf b}^{\delta}),  \nonumber
\end{eqnarray}
where ${\bm \gamma}_t$ and ${\bm \delta}_t$ are the modulation gate and bias based on object $t$. We then modulate all region features by
\begin{eqnarray}
& {\bf r}^n_{tk} = {\bm \gamma}_t \odot  {\bf r}^n_k  + {\bm \delta}_t,  \nonumber
\end{eqnarray}
where $\odot$ is the element-wise multiplication and ${\bf r}^n_{tk}$ means the object-aware region feature in branch $t$. The modulation vectors are expected to emphasize region features containing the object $t$ and weaken unrelated ones. 

In branch $t$, we then conduct cross-modal matching between region features ${\bf r}^n_{tk}$ and the object feature ${\bf o}_{t}$ by
\begin{eqnarray}
&{d}^n_{tk} = {\bf w}^{\top} {\rm tanh}({\bf W}^{c} [{\bf r}^n_{tk}; {\bf o}_{t}; {\bf r}^n_{tk} \odot {\bf o}_{t}; {\bf r}^n_{tk} -{\bf o}_{t}]+ {\bf b}^{c}),  \nonumber
\end{eqnarray}
where ${d}^n_{tk} $ is the matching score of region $k$ in frame $n$ on branch $t$. Next, we apply the softmax function on ${d}^n_{tk}$ to obtain the matching score distribution over regions, given by ${\hat d}^n_{tk} = \frac{{\rm exp}({d}^n_{tk})}{\sum_{k=1}^{K} {\rm exp}({d}^n_{tk})}$. It is used to multi-branch relation reasoning for critical object relation discovery and is also applied to construct the diversity loss between multiple branches.

\subsubsection{Multi-Branch Relation Reasoning}
Next, we develop the multi-branch relation reasoning in each frame to capture critical object relationships by integrating auxiliary branches into the main branch. 
Concretely, the branch $t$ focuses on its corresponding object $t$ and has learnt the object-aware region feature ${\bf r}^n_{tk}$ with the matching score ${\hat d}^n_{tk}$. To build the relation between the main object (i.e. object $1$) and auxiliary object $t$ in each frame, we absorb crucial clues of notable regions from the branch $t$ into the main branch. We estimate the attention weight between the $k$-th region ${\bf r}^n_{1k}$ in the main branch (i.e. branch $1$) and the $l$-th region ${\bf r}^n_{tl}$ in the auxiliary branch $t$ by
\begin{eqnarray}
& \epsilon_{1k,tl}^n = {\bf w}^{\top}{\rm tanh}({\bf W}_1^m {\bf r}^n_{1k} + {\bf W}_2^m {\bf r}^n_{tl} + {\bf W}_3^m{\bf b}^n_{1k,tl} + {\bf b}^m), \nonumber
\end{eqnarray}
where ${\bf b}^n_{1k,tl}$= $[x^n_{1k,tl};y^n_{1k,tl};w^n_{1k,tl};h^n_{1k,tl}]$  is the relative geometry vector between region ${\bf r}^n_{1k}$ and ${\bf r}^n_{tl}$, given by
\begin{eqnarray}
& x^n_{1k,tl} = (x^n_{1k} - x^n_{tl})/w^n_{tl}, \ y^n_{1k,tl} = (y^n_{1k} - y^n_{tl})/h^n_{tl},  \nonumber\\
& w^n_{1k,tl} = {\rm log}(w^n_{1k} / w^n_{tl}), \ h^n_{1k,tl} = {\rm log}(h^n_{1k} / h^n_{tl}).  \nonumber
\end{eqnarray}
Thus, the attention weight $\epsilon_{1k,tl}^n$ is built on object-aware features ${\bf r}^n_{1k}$ and ${\bf r}^n_{tl}$ with the spatial location information. After it, we aggregate the regions relevant to object $t$ from the auxiliary branch $t$ by
\begin{eqnarray}
& {\hat \epsilon}_{1k,tl}^n = \frac{{\rm exp}(\epsilon_{1k,tl}^n)}{\sum_{l=1}^{K} {\rm exp}(\epsilon_{1k,tl}^n)}, \ 
{\bf r}^n_{1k, t} = \sum \limits_{l=1}^{K} {\hat d}^n_{1k} \cdot {\hat d}^n_{tl}  \cdot {\hat \epsilon}_{1k,tl}^n \cdot {\bf r}^n_{tl}, \nonumber
\end{eqnarray}
where ${\bf r}^n_{1k, t}$ is the aggregation feature from branch $t$ for the region $k$ in the main branch.
We first apply the softmax on the attention weights $\epsilon_{1k,tl}^n$ and then aggregate region features with the matching scores ${\hat d}^n_{1k}$ and ${\hat d}^n_{tl}$ as weighting terms. Thus, the relation reasoning between the main object and auxiliary object $t$ will focus on notable regions with higher matching scores ${\hat d}^n_{tl}$ and filter out inessential ones. Simultaneously, by the prior confidence ${\hat d}^n_{1k}$ of the main object, we can enhance the relation modeling for these regions with higher ${\hat d}^n_{1k}$ in the main branch.

After multi-branch relation reasoning from all auxiliary branches, we learn the multi-branch aggregation features $\{ {\bf r}^n_{1k, t} \}_{t=2}^T$ for each region $k$ of frame $n$ in the main branch. We then obtain the final object-aware multi-branch features $\{ \{ {\bf \widetilde r}^n_{k} \}_{k=1}^K \}_{n=1}^N$ by
\begin{eqnarray}
& {\bf \widetilde r}^n_{k} = {\rm ReLU}({\bf r}^n_{1k} + \sum \limits_{t=2}^{T} {\bf r}^n_{1k, t}).  \nonumber
\end{eqnarray}

\subsubsection{Diversity Loss Between Branches}
Considering each branch should only focus on its corresponding object, we devise a diversity loss to make different branches have diverse score distributions over regions. Specifically, we denote the score distribution in the frame $n$ on branch $t$ as ${\bf \hat d}^n_{t}=[{\hat d}^n_{t1},\dots, {\hat d}^n_{tK}]^{\top}$ and calculate the diversity loss by the distribution similarity between multiple branches as follows:
\begin{eqnarray}
& {\mathcal L}_{d}  = \frac{1}{Z}\sum_{n \in {\mathcal S}_{gt}}\sum_{i=1}^{T-1}\sum_{j=i+1}^{T} ({\bf \hat d}^n_{i})^{\top}  ({\bf \hat d}^n_{j}), \nonumber
\end{eqnarray}
where ${\mathcal S}_{gt}$ is the set of frames in the ground truth segment and $Z = \frac{1}{2}|{\mathcal S}_{gt}|T(T-1)$ is the normalization factor. This diversity loss encourages each branch to pay more attention to the region matching with the corresponding object.

\begin{table*}[t]
\centering
\begin{tabular}{c|cccc|cccc}
\hline
\multirow{2}{*}{Method}& \multicolumn{4}{c|}{Declarative Sentence Grounding} & \multicolumn{4}{c}{Interrogative Sentence Grounding} \\
 & m\_tIoU & m\_vIoU&   vIoU@0.3 &vIoU@0.5 &m\_tIoU & m\_vIoU&   vIoU@0.3 &vIoU@0.5  \\
\hline
GroundeR + TALL&\multirow{3}{*}{34.63\%}&9.78\%&11.04\%&4.09\%&\multirow{3}{*}{33.73\%}&9.32\%&11.39\%&3.24\% \\
STPR + TALL&&10.40\%&12.38\%&4.27\%&&9.98\%&11.74\%&4.36\% \\
WSSTG + TALL&&11.36\%&14.63\%&5.91\%&&10.65\%&13.90\%&5.32\%\\
\hline
GroundeR + L-Net&\multirow{3}{*}{40.86\%}&11.89\%&15.32\%&5.45\%&\multirow{3}{*}{39.79\%}&11.05\%&14.28\%&5.11\% \\
STPR + L-Net&&12.93\%&16.27\%&5.68\%&&11.94\%&14.73\%&5.27\%\\
WSSTG + L-Net&&14.45\%&18.00\%&7.89\%&&13.36\%&17.39\%&7.06\%\\
\hline   
STGRN&48.47\%&19.75\%&25.77\%&14.60\%&46.98\%&18.32\%&21.10\%&12.83\%\\
\hline
OMRN (Ours)&{\bf 50.73\%}&{\bf 23.11\%}&{\bf 32.61\%}&{\bf 16.42\%}&{\bf 49.19\%}&{\bf 20.63\%}&{\bf 28.35\%}&{\bf 14.11\%}\\
\hline
\end{tabular}
\caption{Performance Evaluation Results on the VidSTG Dataset.}\label{table:mainexp}
\end{table*}

\subsection{Spatio-Temporal Localization}
In this section, we apply a spatio-temporal localizer to predict the spatio-temporal tube of the queried object based on final region features $\{ \{ {\bf \widetilde r}^n_{k} \}_{k=1}^K \}_{n=1}^N$, including the temporal moment localization and spatial region localization.

For spatial localization, we estimate the region confidence scores according to the main object feature ${\bf o}_1$  by
\begin{eqnarray}
& {p}^{n}_k = \sigma( ({\bf W}^r {\bf \widetilde r}^n_{k})^{\top}  ({\bf W}^o {\bf o}_1)), \nonumber
\end{eqnarray}
where $\sigma$ is the sigmoid function and ${p}^{n}_k$ is the confidence score of region $k$ in frame $n$.
Next, we apply the spatial loss to guide the spatial localization where we only consider frames in the set ${\mathcal S}_{gt}$, i.e. in the ground truth segment.
We first compute the IoU score ${\rm IoU}^{n}_k$ of each region with the corresponding ground truth region and then calculate the spatial loss by
\begin{equation}
\begin{aligned}
{\mathcal SL}^n_{k}  &= (1 - {\rm IoU}_{k}^n) \cdot {\rm log}(1 - {p}_{k}^n) + {\rm IoU}_{k}^n \cdot {\rm log}({p}_{k}^n), \\ \nonumber
{\mathcal L}_{s} &= -\frac{1}{|{\mathcal S}_{gt}|K}\sum_{n \in {\mathcal S}_{gt}}\sum_{k=1}^{K}  {\mathcal SL}^n_{k}.
\end{aligned}
\end{equation}

For temporal localization, we sample a set of candidate segments and estimate their confidence scores with the boundary adjustment. Concretely, we first adopt a spatial attention to aggregate the final region features by
\begin{eqnarray}
&\zeta^n_{k} = {\bf w}^{\top}{\rm tanh}({\bf W}^{f}_{1} {\bf \widetilde r}_{k}^{n}+ {\bf W}^{f}_{2}{\bf o}_{1}+{\bf b}^{f}), \nonumber \\
&{\bf f}^{n} = \sum_{k=1}^{K} {\rm softmax}({\zeta^n_{k}}) \cdot {\bf \widetilde r}_{k}^{n}, \nonumber
\end{eqnarray}
where ${\bf f}^n$ is the object-aware feature of frame $n$. We then input $\{{\bf f}^n\}_{n=1}^N$ into a Bi-GRU to learn context features $\{{\bf \widetilde f}^n\}_{n=1}^N$. Next, we regard each frame as a sample center and define $H$ candidate segments with widths $\{{w}^h\}_{h=1}^H$ at each center. 
After it, we generate the confidence scores and their boundary offsets by 
\begin{eqnarray}
& {\bf c}^{n} = \sigma({\bf W}^c {\bf \widetilde f}^n + {\bf b}^c), \  {\bf l}^{n} = {\bf W}^l {\bf \widetilde f}^n+ {\bf b}^l, \nonumber
\end{eqnarray}
where ${\bf c}^{n} \in \mathbb{R}^{H}$ represent confidence scores of $H$ candidates at step $n$ and ${\bf l}^{n} \in \mathbb{R}^{2H}$ are their offsets. Similar to the spatial loss, we apply the temporal alignment loss for segment selection, which is based on the temporal IoU score ${\rm tIoU}_{h}^n$ of each candidate segment with the ground truth segment, given by
\begin{equation}
\begin{aligned}
{\mathcal TL}^n_{h}  &= (1 - {\rm tIoU}_{h}^n) \cdot {\rm log} (1 - {c}_{h}^n) + {\rm tIoU}_{h}^n \cdot {\rm log}({c}_{h}^n), \\ \nonumber
{\mathcal L}_{t} &=  -\frac{1}{NH}\sum_{n=1}^{N}\sum_{h=1}^{H}  {\mathcal TL}^n_{h}.
\end{aligned}
\end{equation}

Next, we select the segment with the highest ${c}_{h}^n$ and adjust its temporal boundaries by the offset $(l_s,l_e)$ from  ${\bf l}^{n}_{h}$. Here we develop another regression loss to train the offsets. With the original boundaries $({s}, {e})$ of the selected segment and ground truth $({\hat s}, {\hat e})$, we first compute the ground truth offsets ${\hat l}_s = s - {\hat s}$ and ${\hat l}_e = e - {\hat e}$ and compute the regression loss by
\begin{eqnarray}
& {\mathcal L}_{r} = {\rm R}({l}_s - {\hat l}_s) + {\rm R}({l}_e - {\hat l}_e), \nonumber
\end{eqnarray}
where ${\rm R}$ is the smooth L1 function.

Eventually, we apply the multi-task loss to train our OMRN method in an end-to-end manner, given by
\begin{eqnarray}
& {\mathcal L}_{OMRN} = {\lambda}_1 {\mathcal L}_{s} +  {\lambda}_2 {\mathcal L}_{t} + {\lambda}_3 {\mathcal L}_{r} + {\lambda}_4 {\mathcal L}_{d}, \nonumber
\end{eqnarray}
where ${\lambda}_1$, ${\lambda}_2$, ${\lambda}_3$ and ${\lambda}_4$ control the balance of four losses.

During inference, we first detect the segment with the highest temporal confidence score, fine-tune its boundaries by its offsets and then select the regions with the highest spatial scores within the selected segment to generate the target tube.

\section{Experiments}

\subsection{Experimental Settings}
\subsubsection{Dataset} 
We conduct experiments on a large-scale spatio-temporal video grounding dataset VidSTG~\cite{zhang2020does}, which is constructed from the video object relation dataset VidOR~\cite{shang2019annotating} by annotating the natural language descriptions. As we know, VidSTG is the only grounding dataset on unaligned video-sentence data and multi-form sentences. 
Specifically, VidSTG contains 5,563, 618 and 743 videos in the training, validation and testing sets, totaling 6,924 videos. There are 99,943 sentence annotations for 80 types of queried objects, including 44,808 and 55,135 for declarative and interrogative sentences, respectively. 
Moreover, the duration of the videos is 28.01s and temporal tube length is 9.68s on average. And declarative and interrogative sentences have about 11.12 and 8.98 words, respectively.

\subsubsection{Implementation Details} 
During data preprocessing, we extract 1,024-d region features by a pre-trained Faster R-CNN~\cite{ren2015faster}. We sample 5 frames per second and extract $K=20$ regions for each frame. For language, we apply a pre-trained Glove embedding to obtain 300-d word features and use the NLTK to recognize the nouns in sentences.
As for modeling setting, we set $\alpha$ to 0.6, $L$ to 5 and set $\lambda_1$,  $\lambda_2$,  $\lambda_3$ and $\lambda_4$ to 1.0, 1.0, 0.001 and 1.0, respectively. We define $H=9$ candidate segments at each step with temporal widths $[3,9,17,33,65,97,129,165,197]$. We set the dimensions of all projection matrices and biases to 256 and set the hidden state of each direction in BiGRU to 128. We employ an Adam optimizer with the initial learning rate 0.0005.

\subsubsection{Evaluation Criteria}
We apply the criterion m\_tIoU to evaluate the temporal grounding performance and use m\_vIoU and vIoU@R to estimate the spatio-temporal accuracy as \cite{zhang2020does}. Concretely, the m\_tIoU is the average temporal IoU of selected segments with the ground truth. We define vIoU as the spatio-temporal IoU between the predicted and ground truth tubes, given by $ {\rm vIoU} = \frac{1}{|{\mathcal S}_{p} \cup {\mathcal S}_{gt}|} \sum_{n \in {\mathcal S}_{p} \cap {\mathcal S}_{gt}} {\rm IoU}(r^n, {\hat r}^n)$. The ${\mathcal S}_{p}$ is the frame set in the predicted segment and ${\mathcal S}_{gt}$ is the frame set in the ground truth. The $r^n$, ${\hat r}^n$ are the predicted and ground truth regions in frame $n$. The m\_vIoU is the mean vIoU of all test samples and vIoU@R is the rate of testing samples with vIoU $>$ R.

\subsection{Performance Comparison}
As an emerging task, only the STGRN method~\cite{zhang2020does} is designed for STVG on unaligned data. Besides it, we combine the temporal grounding methods TALL~\cite{gao2017tall} and L-Net~\cite{chen2019localizing} with spatio-temporal grounding approaches on aligned data as the baselines.
Specifically, TALL employs a proposal-and-selection framework for the temporal localization and L-Net develops frame-by-word interactions for holistic segment selection. 
Given the predicted segment, GroundeR~\cite{rohrbach2016grounding} is a visual grounding method to retrieve the target region in each frame. And the STPR~\cite{yamaguchi2017spatio} and WSSTG~\cite{chen2019weakly} approaches apply tube pre-generation in the segment and then rank these tubes by cross-modal estimation. 
Concretely, the original STPR is applied to multi-video person grounding, we extend it to single-video grounding for diverse objects. The WSSTG originally use a weakly-supervised ranking loss but we replace it with a supervised triplet loss~\cite{yang2019cross}. Thus, there are 6 combinations such as WSSTG+TALL and STPR+L-Net.

The overall experiment results are shown in Table~\ref{table:mainexp} and we can find some interesting points:
\begin{itemize}
\item On the whole, the grounding performance of all models for interrogative sentences is lower than for declarative sentences, validating the unknown objects without explicit characteristics are more difficult to ground.
\item For temporal grounding, region-level methods STGRN and OMRN outperform the frame-level methods TALL and L-Net, which demonstrates the fine-grained region modeling is beneficial to determine the accurate temporal boundaries of target tubes.
\item For spatio-temporal grounding, the GroundR+$\{\cdot\}$ approaches ignore the temporal dynamics of objects and achieve the worst performance, suggesting it is crucial to capture the object dynamics among frames for high-quality spatio-temporal video grounding.
\item In all criteria, our OMRN achieves the remarkable performance improvements compared with baselines. This fact shows our method can effectively focus on the notable regions by object-aware multi-branch region modeling with the diversity loss and capture critical object relations by multi-branch reasoning.
\end{itemize}
Furthermore, given the temporal ground truth segment during inference, we compare the spatial grounding ability of our OMRN method with baselines. The results are shown in Table~\ref{table:temgt} and we do not separate declarative and interrogative sentences here. We can see that our OMRN still achieves the apparent performance improvement on all criteria, especially for vIoU@0.3. This demonstrates our OMRN approach is still effective when applied to aligned segment-sentence data.

\begin{table}[t]
\centering
\scalebox{0.95}{
\begin{tabular}{c|ccc}
\hline
Methods &   m\_vIoU &vIoU@0.3 &vIoU@0.5\\
\hline
GroundeR + Tem. GT& 27.31\%&40.53\%&20.86\%\\
STPR + Tem. GT&28.20\%&42.08\%&21.75\%\\
WSSTG + Tem. GT&31.51\%&46.99\%&27.65\%\\
STGRN + Tem. GT& 36.75\%&50.78\%& 32.93\%\\
\hline
OMRN + Tem. GT& {\bf 39.57\%}&{\bf 58.19\%}&{\bf 37.91\%}\\
\hline
\end{tabular}
}
\caption{Evaluation Results with the Temporal Ground Truth.}\label{table:temgt}
\end{table}

\begin{table}[t]
\centering
\scalebox{0.95}{
\begin{tabular}{c|cccc}
\hline
Methods &  m\_tIoU& m\_vIoU &vIoU@0.3 &vIoU@0.5\\
\hline
w/o. OM&47.78\%&19.85\%&28.55\%&12.99\%\\
w/o. DL&48.32\%&20.08\%&28.80\%&13.42\%\\
w/o. CM&47.23\%&19.06\%&27.08\%&12.25\%\\
\hline
w/o. TA&48.92\%&20.50\%&29.76\%&13.61\%\\
w/o. CA&48.85\%&20.73\%&29.87\%&14.02\%\\
\hline
full&{\bf 49.88\%}&{\bf 21.73\%}&{\bf 30.26\%}&{\bf 15.14\%}\\
\hline
\end{tabular}
}
\caption{Ablation Results on the VidSTG Dataset.}\label{table:ablation}
\end{table}

\subsection{Ablation Study}
We next verify the contribution of each part of our method by ablation study. We remove one key component at a time to generate an ablation model. The object-aware multi-branch modeling is vital in our method, so we first remove the object-aware modulation from each branch as \textbf{w/o. OM}. We then discard the diversity loss from the multi-task loss, denoted by \textbf{w/o. DL}. Further, we remove the cross-modal matching from all branches and discard the weighting terms ${\hat d}^n_{1k}$ and ${\hat d}^n_{tl}$ in multi-branch relation reasoning, denoted by \textbf{w/o. CM}. Note that in this ablation model, the diversity loss is also ineffective due to the lack the matching score distributions. Next, we develop the ablation study on the basic region and object modeling. We discard the temporal region aggregation from region modeling as \textbf{w/o. TA} and remove the context attention during object extraction as \textbf{w/o. CA}.

The ablation results are shown in Table~\ref{table:ablation}. We can find all ablation models have the performance degradation compared with the full model, showing each above component is helpful to improve the grounding accuracy.
And the ablation models \textbf{w/o. OM}, \textbf{w/o. DL} and \textbf{w/o. CM} have the lower accuracy than \textbf{w/o. TA} and \textbf{w/o. CA}, which suggests the object-aware multi-branch relation reasoning plays a crucial role in high-quality spatio-temporal grounding.
Moreover, the model \textbf{w/o. CM} achieves the worst performance, validating the cross-modal matching with the diversity regularization is very important to incorporate language-relevant region features from auxiliary branches into the main branch.

\begin{figure}[t]
\centering
\includegraphics[width=0.48\textwidth]{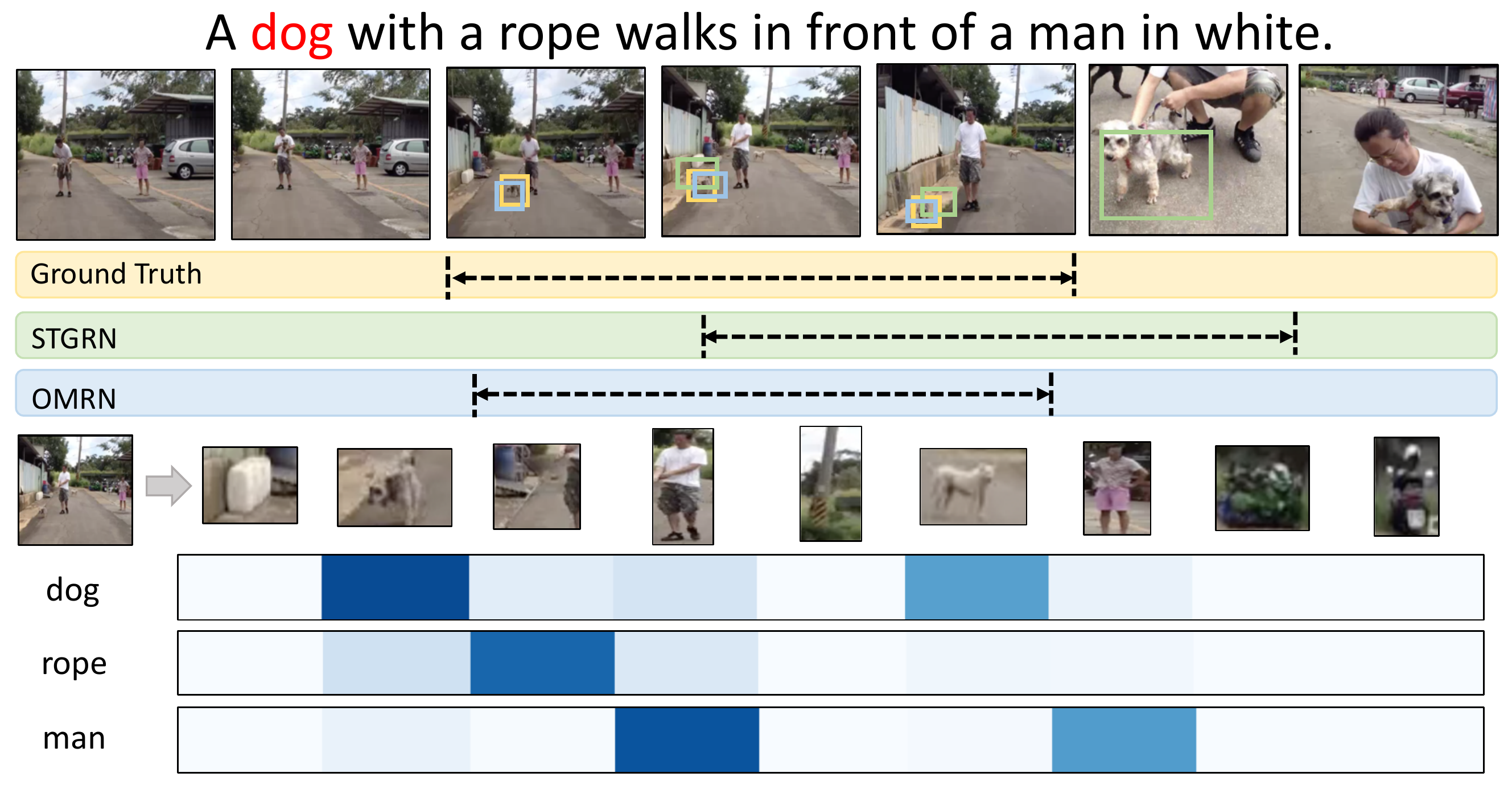}
\caption{A typical example of the grounding result.}\label{fig:atten}
\end{figure}

\subsection{Qualitative Analysis}
To qualitatively validate the effectiveness of our OMRN method, we display a typical example in Figure~\ref{fig:atten}.
The sentence describes a short-term state of the "dog" and requires to capture object-aware fine-grained relations. By intuitive comparison, our OMRN can retrieve the more accurate temporal boundaries and spatio-temporal tube of the "dog" than the best baseline STGRN. Furthermore, we display the object-region matching score distribution in the example, where we visualize the matching scores between three objects (i.e. "dog", "rope" and "man") and the regions of the 4-th frame. Although there are a woman and another dog in the frame, our method can still eliminate the interference and focus on the notable region containing the corresponding object, e.g., the object "dog" assigns a higher score to the 2-th region rather than the 6-th region.

\section{Conclusion}
In this paper, we propose a novel object-aware multi-branch relation network for STVG. The method can effectively focus on the vital regions by object-aware multi-branch region modeling and capture sufficient object relations by multi-branch reasoning for high-quality spatio-temporal grounding.

\section*{Acknowledgments}
This work was supported by the National Key R\&D Program of China under Grant No. 2018AAA0100603, Zhejiang Natural Science Foundation LR19F020006 and the National Natural Science Foundation of China under Grant No.61836002, No.U1611461 and No.61751209.
This research is partially supported by the Language and Speech Innovation Lab of HUAWEI Cloud.

\clearpage

\bibliographystyle{named}
\bibliography{ref}

\end{document}